\documentclass{sisl-tufte}  

\usepackage{standalone}
\usepackage{pgfplots} 
\usepackage[noend]{algcompatible}

\usepgfplotslibrary{groupplots}
\usetikzlibrary{arrows.meta}
\usetikzlibrary{arrows}
\usetikzlibrary{positioning}
\usetikzlibrary{shapes}
\usetikzlibrary{backgrounds}
\tikzset{>=stealth'}
\pgfplotsset{compat=newest}
\pgfplotsset{every axis legend/.append style={%
		cells={anchor=west},
		label style={font=\Huge}}
}
\usepgfplotslibrary{polar}

\begin{document}

\title{Decomposition Methods with Deep Corrections \\ for Reinforcement Learning}

\author{\name Maxime Bouton \email boutonm@stanford.edu \\
        \name Kyle D. Julian  \email kjulian3@stanford.edu \\   
        \name Alireza Nakhaei \email anakhaei@honda-ri.com \\
        \name Kikuo Fujimura \email  kfujimura@honda-ri.com \\
        \name Mykel J. Kochenderfer \email mykel@stanford.edu\\
        }

\maketitle

\begin{abstract}  
Decomposition methods have been proposed to approximate solutions to large sequential decision making problems.
In contexts where an agent interacts with multiple entities, utility decomposition can be used to separate the global objective into local tasks considering each individual entity independently. 
An arbitrator is then responsible for combining the individual utilities and selecting an action in real time to solve the global problem. Although these techniques can perform well empirically, they rely on strong assumptions of independence between the local tasks and sacrifice the optimality of the global solution. 
This paper proposes an approach that improves upon such approximate solutions by learning a correction term represented by a neural network. 
We demonstrate this approach on a fisheries management problem where multiple boats must coordinate to maximize their catch over time as well as on a pedestrian avoidance problem for autonomous driving.
In each problem, decomposition methods can scale to multiple boats or pedestrians by using strategies involving one entity.
We verify empirically that the proposed correction method significantly improves the decomposition method and outperforms a policy trained on the full scale problem without utility decomposition.


\end{abstract}


\section{Introduction}

Handcrafting decision making strategies for autonomous systems in complex environments is difficult. The burden is placed on the designer to anticipate the wide variety of possible situations and explicitly program how the autonomous agent should behave. There has been growing interest in automatically deriving suitable behavior by optimizing decision policies using reinforcement learning (RL)~\cite{dmu}. An agent interacts with the environment (through simulation or collected data) and acts  based on its observations to maximize expected long term return. 

There are many domains where the agent must interact with multiple entities. For example, in the autonomous driving context, other entities may include  vehicles on the road that must be avoided. In resource allocation problems~\cite{russel2003,tesauro2005}, all the agents must be coordinated and assigned to some resources to maximize a common objective. We will refer to these two categories of problems as \textit{multiple entities} when one agent must react against non-controlled entities (\textit{e.g.} collision avoidance) and as \textit{multi-agent} when several agents must cooperate or be coordinated (\textit{e.g.} resource allocation).
RL can have difficulty scaling to such domains because the state space grows exponentially with the number of entities and agents. The general problem of decentralized control of multiple agents in a stochastic, partially observed environment has been shown to be intractable~\cite{bernstein2000}. One way to approximate solutions to problems with multiple entities is to decompose the state space into pairwise interactions and learn the utility function with each entity individually. In multi-agent problems, each agent is considered independently. The solutions to the individual subproblems can be combined in real-time through an arbitrator that maximizes the expected utility \cite{rosenblatt2000}. Applications to aircraft collision avoidance with multiple intruders explore summing or using the minimum state-action values \cite{chryssanthacopoulos2012,ong2015}. While these approaches tend to perform well empirically, they sacrifice optimality. The solution to each subproblem assumes that its individual policy will be followed regardless of other entities, in contrast to the global policy considering all subproblems~\cite{russel2003}. 

Previous approaches to scale decision algorithms using decomposition methods relied on a distributed agent architecture~\cite{rosenblatt2000}. Each agent is responsible for addressing one of the multiple objectives required to achieve a complex task. At each time step an arbitrator must decide between the different actions recommended by these agents and address possible conflicts. Possible arbitration strategies are command fusion~\cite{chryssanthacopoulos2012}, voting~\cite{rosenblatt2000}, lexicographic ordering~\cite{wray2017} or utility fusion~\cite{rosenblatt2000, russel2003}. It has been shown that utility fusion offers a more principled way of deciding between the individual agents compared to a voting-based approach or command~fusion~\cite{rosenblatt2000}. An alternative approach in leader follower scenarios consists of reducing the problem of controlling the group of agent to controlling a single group leader~\cite{hung2017}. Other approaches rely on distributed learning algorithms, such as independent Q-learning, where each agent learns a policy without being aware of the other agents actions~\cite{tan1993}. The underlying assumption of independence between agents in these algorithms trades off the benefit of collaboration for an easier learning of the task~\cite{claus1998}.

Utility decomposition methods have been used in many practical applications in both adversarial and cooperative settings. \citeauthor{chryssanthacopoulos2012} applied utility decomposition to a collision avoidance problem where an aircraft must avoid multiple other aircraft~\cite{chryssanthacopoulos2012}. A similar approach is proposed for avoiding potential debris in the airspace~\cite{tompa2018}. Van \citeauthor{vanderpol2016} used utility decomposition to coordinate traffic lights across a road network~\cite{vanderpol2016}. Instead of considering each cooperating agent as an individual, one can formulate the problem through pairwise interactions, as demonstrated on a wildfire surveillance problem~\cite{julian2018}. In the cooperative case, the arbitrator selects a joint action. Efficient search strategies might be required to compute the joint action maximizing the sum of the individual utilities~\cite{oliehoek2013, ong2015}. 
 In settings where the reward function can be additively decomposed, an arbitrator maximizing the sum of the individual utilities can be optimal under certain conditions~\cite{russel2003}. However, this arbitration method will be suboptimal for the global task in many problems. In  resource allocation tasks, the individual agents might have selfish policies, so arbitrating selfish policies by maximizing the sum of utilities can lead to a problem known as ``the tragedy of the commons''~\cite{russel2003}. To address the problem of selecting an appropriate utility fusion function, it has been suggested to learn the structure of the decomposition using a neural network representation~\cite{sunehag2018}.
 

Another approach to scaling reinforcement learning algorithms leverages the representational power of deep neural networks in modeling the utility function. Although this approach can handle domains such as Atari games and complicated manipulation tasks, it often requires millions of training examples and long training times \cite{mnih2015, gu2017}. To minimize training time, transfer learning can take advantage of prior knowledge. For example, an existing controller or human expert can gather high reward demonstrations to train an autonomous agent \cite{smart2002}. Another transfer learning approach uses a regularizing term to guide the agent towards regions where it has prior knowledge of the appropriate behavior \cite{gottwald2017}. Multi-fidelity optimization has been combined with model-based reinforcement learning \cite{cutler2015}, where a learning framework is proposed for efficiently deciding between sampling from an inexpensive low-fidelity simulator and real world data. 

This paper presents an approach for scaling RL algorithms to problems with multiple entities or multiple agents using utility decomposition with corrections represented by a deep neural network. Borrowing concepts from multi-fidelity optimization, a correction model is trained to improve an existing pairwise approximation of the optimal utility function.
We first present how approximate solutions can be found by decomposing sequential decision problems into individual subproblems. The value functions to the subproblems can be combined using utility fusion to approximate the utility function of the global task. 
We then introduce the concept of surrogate corrections with neural representation, originally from multi-fidelity optimization, to improve this globally suboptimal policy. Finally, we empirically demonstrate the advantage of this technique on an autonomous driving scenario and a fisheries management problem. It is shown that the policy resulting from the correction method is faster to learn and outperforms a policy learned~directly.

\section{Background} 

This section introduces a common mathematical framework for formulating sequential decision making problems as an optimization of a reward over time. A utility function is defined to represent the expected reward that the agent can accumulate from a given state while acting optimally. For large problems, the utility function can be approximated by a combination of utility functions of simpler sub-problems.

\subsection{Markov Decision Processes}
Sequential decision problems are commonly formulated as Markov decision processes (MDPs).
An MDP can be formally defined by the tuple $(\mathcal{S}, \mathcal{A}, T, R, \gamma)$, with state space $\mathcal{S}$, action space $\mathcal{A}$, state transition function $T$, reward function $R$, and discount factor $\gamma$. 
At time $t$, an agent chooses an action $a_t\in\mathcal{A}$ based on observing state $s_t\in \mathcal{S}$. The agent then receives a reward $r_t = R(s_t, a_t)$.
At time $t+1$, the environment transitions from $s_t$ to a state $s_{t+1}$ with  probability $\Pr(s_{t+1}\mid s_t, a_t) = T(s_t, a_t, s_{t+1})$. The agent's objective is to maximize the accumulated expected discounted reward given by $\sum_{t=0}^{\infty} \gamma^t r_t$. 

A policy $\pi : \mathcal{S} \rightarrow \mathcal{A}$ defines what action to execute at a given state. Each policy can be associated to a state-action value function $Q^\pi : \mathcal{S}\times\mathcal{A} \rightarrow \mathbb{R}$, representing the expected discounted value of following the policy $\pi$. The optimal state value function of an MDP satisfies the Bellman equation:
\begin{equation}
	Q^*(s, a) = R(s, a) + \gamma\sum_{s'}T(s, a, s')\max_{a'} Q^*(s', a')  
\end{equation}
where $s$ is the current state of the environment and $s'$ a next state reachable by taking action $a$.
This paper focuses on approaches where the transition function is not directly available. Instead, the agent has access to a generative model from which the next state is sampled. The Bellman equation can be defined more generally as an expectation over the next state:
\begin{equation}
	Q^*(s, a) = \mathbb{E}_{s'} [ R(s, a) + \gamma\max_{a'} Q^*(s', a') ] 
	\label{eq:bellman}
\end{equation}
Once $Q^*$ is computed, the associated optimal policy is given by taking the action maximizing the value at a given state: 
\begin{equation}
    \pi^*(s) = \argmax_a Q^*(s,a)
\end{equation}
Similarly, we define the utility of a given state as follows:
\begin{equation}
	U^*(s) = \max_a Q^*(s,a)
\end{equation}

This paper is concerned with single agent planning problems against multiple non-cooperative entities as well as multi-agent planning problems. In the multi-agent setting, the action space consists of the joint action space for all agents to control. \Cref{sec:fisheries} will discuss an application to a multi-agent coordination problem and \cref{sec:crosswalk} is dedicated to a case study of a single agent problem.

\subsection{Utility Decomposition}\label{sec:fusion}

Utility decomposition, which is sometimes called Q-decomposition, involves combining the utility functions associated with simple decision making tasks to approximate the solution to a more complex task~\cite{russel2003}. We will refer to the complex task as the global problem and the subtasks as local problems. It is often the case that the state space of the local problems is a subset of the state space of the global problem. Each local problem $i$ is formulated as an MDP and is first solved in isolation. The function $Q^*_i$ represents the optimal value function to solve the subtask $i$. In non-cooperative multiple entities settings, these local problems are pairwise interactions between the decision agent and one of the entities to act against. In cooperative multiple agent settings, the local problems are instances of the global problem with a single agent or a subset of agents. \citeauthor{russel2003} suggest a decomposition based on the different additive terms in the reward function of the global problem~\cite{russel2003}. Solving the global task is then achieved by fusing the utilities associated with each local task. More formally, utility fusion requires defining a function $f$ such that: 
\begin{equation}
Q^*(s, a) \approx f(Q^*_1(s_1, a), \ldots, Q^*_n(s_n, a))
\label{eq:fusion}
\end{equation}
where $Q^*$ is the optimal value function associated with the global task. The state variable can also be decomposed, and we can assume that each of the value functions $Q^*_i$ uses a subset of the information contained in $s$ to solve the simpler subtask. One approach to this decomposition involves decomposing the reward function additively~\cite{russel2003}. Each term of the sum is then optimized individually by a sub-agent. To solve the global problem they choose $f$ to be the sum of the individual value functions. They show that the solution does not achieve optimality using the tabular Q-learning algorithm, although decomposition of the reward model has been demonstrated to work well empirically \cite{rosenblatt2000}.

The simplicity of the approach makes it a very appealing technique to find an approximately optimal solution to a complex decision making problem. In many problems involving non-cooperative multiple agents, utility fusion can help to scale the solution. A common setting for utility decomposition is when an autonomous agent must make decisions involving multiple entities. One example is collision avoidance problems \cite{chryssanthacopoulos2012}, where the autonomous agent must avoid multiple moving targets. In this setting, the decomposition approach consists in solving the subtask of avoiding a single target. When interacting with multiple entities, the agent computes the value associated with avoiding each entity $i$ assuming it is the only entity to avoid. Solving for pairwise interactions rather than the global problem requires exponentially fewer computations since the size of the state space often grows exponentially with the number of entities. The global utility is then computed by summing the individual value functions or by taking the minimum over each entity to avoid. Summing the value functions as follows:
\begin{equation}
	Q^*(s, a) \approx \sum_i Q^*_i(s_i, a) \label{eq:sum}\\
\end{equation}
implies that all targets to avoid are independent.
\Cref{eq:sum} equally weighs the utility of a user in a safe zone of the environment and a user in a more dangerous zone. Instead, another strategy is to take the minimum as follows:
\begin{equation}
	Q^*(s, a) \approx \min_i Q^*_i(s_i, a) \label{eq:min}
\end{equation}
In \cref{eq:min}, taking the minimum will consider the target with the lowest utility. Given a reward function penalizing for collisions, it will consider the target that action $a$ is most likely to harm. This approach is more risk averse. A detailed example of the decomposition method for a collision avoidance problem for autonomous driving is discussed in \cref{sec:crosswalk}. \Cref{eq:sum,eq:min} will be referred to as the max-sum and max-min approaches, respectively, because the action to execute is obtained by $\argmax_a f(Q^*_1(s_1, a), \cdots, Q^*_n(s_n, a))$. Decomposition is also helpful in problems where several agents must be coordinated. Again one can consider each agent individually and find a joint action by maximizing the sum of the individual utilities. An example of such a scenario is discussed in \cref{sec:fisheries}.

A disadvantage of this decomposition method is that it requires choosing a fusion function that will play the role of arbitrator between the different utilities considered. The choice of this function can greatly affect the performance of the global policy and has no guarantee of optimality. In fact, it can be shown that when the individual utility functions are computed, they converge to ``selfish" estimates \cite{russel2003}. 
Instead this solution provides a low-fidelity approximation of the optimal value function for  little computational cost once the individual utility functions are computed.

\section{Deep Corrections}

A method that leverages the decomposed utility functions to guide an RL algorithm to reach an optimal policy with less experience is introduced here.

\subsection{Deep Q-Networks}\label{sec:qlearning}

In MDPs with discrete state and action spaces, the value function can often be represented by a table and computed using dynamic programming. In many reinforcement learning problems, the state space can be very large or even continuous, making it impossible to explicitly represent the value associated to every possible state. Instead, the value function is represented by a parametric model such as a neural network: $Q(s, a; \theta)$, referred to as a deep Q-network (DQN). This technique approximates the value of every possible state with a limited number of parameters $\theta$.

Computing the parameters of the network is formulated as a learning problem. An objective function to minimize, based on \cref{eq:bellman}, can be expressed as follows: 
\begin{equation}
	J(\theta) = \mathbb{E}_{s'}[(r + \gamma\max_{a'}Q(s', a'; \theta_-) - Q(s, a; \theta))^2]
\end{equation}

The parameter $\theta_-$ defines a fixed target network. The loss function is computed and minimized using sampled experiences. An experience comes from interacting with the environment over one time step where the agent in state $s$ takes action $a$, transitions to state $s'$, and receives reward $r$. The action $a$ is selected using a $\epsilon$-greedy strategy.  The original Q-learning algorithm uses a single Q-network in the objective function, but fixing the parameters in a target network for several steps has been shown to help the convergence of the algorithm \cite{mnih2015}. If $Q$ satisfies the Bellman equation, then the loss should be zero. By taking the gradient of $J$ with respect to the parameters $\theta$, we obtain the following update rule given an experience $(s, a, r, s')$:
\begin{equation}
	\theta \leftarrow \theta + \alpha (r+ \gamma\max_{a'}Q(s', a'; \theta_-) - Q(s, a; \theta))\nabla_\theta Q(s,a;\theta)
\end{equation}
where $\alpha$ is the learning rate, a hyperparameter of the algorithm.

This algorithm can be unstable and difficult to tune. Fortunately, there are several innovations to improve network training, such as double DQN, dueling network architectures, and prioritized experience replay \cite{schaul2016,van2016deep,wang2016}. These are the three improvements to the deep Q-learning algorithm that we used in this paper. 

Although this algorithm can handle high-dimensional state spaces, it requires many experience samples to converge (on the order of millions for Atari games \cite{mnih2015, wang2016}). Experience can sometimes be generated using simulators, but for applications where simulation is expensive or nonexistent, the deep Q-learning algorithm may be impractical. When applying reinforcement learning to accomplish a difficult task, a possibility is to decompose it into simpler subtasks for which you can afford to run algorithms such as deep Q-learning.

\subsection{Policy Correction}\label{sec:correction}

Computing the optimal Q function with the deep Q-learning algorithm requires many experience samples mostly because the agent must go through a random exploration phase. By introducing prior knowledge into the algorithm, the amount of exploration required to converge to a good policy is significantly reduced. In this work, we consider that prior knowledge takes the form of a value function. No specific representation is needed. The decomposition method described above would be a simple and efficient way to generate prior knowledge. There are many techniques to approximate the value function, such as discretizing a continuous state space or using domain expertise such as physics-based models. If a controller already exists, a value function can be estimated using the Monte Carlo evaluation algorithm \cite{sutton1998}. Given an approximately optimal value function, the objective is to leverage this prior knowledge to learn the optimal value function with as little computation as possible.

The method we propose is inspired by multi-fidelity optimization~\cite{eldred2006}. Consider a setting with two models of different fidelities: an expensive high-fidelity model ($f_{\text{hi}}$) and a low-fidelity model ($f_{\text{lo}}$) providing an inexpensive but less accurate approximation. In multi-fidelity optimization, the goal is to maximize an objective function $f$ approximated by these two simulators. However, using only the high-fidelity model would be too expensive since many queries are probably needed to optimize $f$. On the other hand, relying only on the low-fidelity model might lead to a poor solution. Instead, we could construct a parametric model that approximates $f_{\text{hi}}$ but is inexpensive to compute. Such a surrogate model can represent the difference between the high-fidelity and the low-fidelity models~\cite{eldred2006, rajnarayan2008}:
\begin{equation}
f_{\text{hi}}(x) \approx f_{\text{lo}}(x) + \delta(x)
\end{equation}
where $x$ is the design variable and $\delta$ is a surrogate correction learned using a limited number of samples from $f_{\text{hi}}$. Commonly, $\delta$ is modeled using Gaussian processes or parametric models. In this work, we focused on additive surrogate correction, but a multiplicative correction or a weighted combination of both would be a valid approach as~well~\cite{eldred2006}. 

Reinforcement learning can be reshaped into a multi-fidelity optimization problem. Let $Q_{\text{lo}}$ be our low-fidelity approximation of the value function obtained from prior knowledge or from a decomposition method. The high-fidelity model that we want to obtain is the optimal value function. The surrogate correction for reinforcement learning can be described as follows: 
\begin{equation}
	Q^*(s, a) \approx Q_{\text{lo}}(s, a) + \delta(s, a; \theta)
	\label{eq:correction}
\end{equation}
In regular multi-fidelity optimization problems, samples from $f_{\text{hi}}(x)$ are used to fit the correction model. In reinforcement learning, $Q^*(s, a)$ is unknown, so a temporal difference approach is used to derive a learning rule similar to the Q-learning algorithm. We seek to minimize the same loss with the target network as presented in \cref{sec:qlearning}. The main difference lies in the representation of the Q function, which can be substituted by \cref{eq:correction}. Only the corrective part is parameterized. As a consequence, when taking the gradients with respect to the parameters, the update rule becomes:
\begin{equation}\label{eq:gradient}
	\theta \leftarrow \theta - \alpha [R(s, a) + \gamma\max_{a'}(Q_{\text{lo}}(s', a') + \delta(s', a'; \theta_-)) -  Q_{\text{lo}}(s, a) - \delta(s, a; \theta)]\nabla_\theta \delta(s,a,\theta)
\end{equation}
\Cref{eq:gradient} would be the only difference to the algorithm if another correction method is used. We can generalize the notation to any parametric corrective function $f$: $Q(s, a;\theta)\approx f(s, a, Q_{\text{lo}}(s, a) ; \theta)$. As the deep Q Learning algorithm relies on efficient gradient computation, it assumes that the gradient of $f$ with respect to $\theta$ can be computed easily.

\begin{algorithm}
	\caption{Deep Corrections Algorithm}
	\begin{algorithmic}[1]
		\STATE \textbf{input: } $Q_{\text{lo}}$, $G$ a generative model of the environment, an exploration strategy (\textit{e.g.} $\epsilon$ greedy).
        \STATE \textbf{hyperparameters: } similar as DQN \cite{mnih2015}, let $T$ be the target network update frequency
        \STATE Initialize $\delta$
		\FOR{each episode}
        	\STATE initialize state $s$
            \WHILE{$s$ is not a terminal state}
            	\STATE select $a =  \begin{cases}
                					\text{random with probability } \epsilon \\
                                    \argmax_{a'} (Q_{\text{lo}}(s, a') + \delta(s, a'; \theta))
                                    \end{cases}$
                \STATE $s', r \sim G(s, a)$  
                \STATE Store transition $(s, a, r, s')$ in a replay buffer
                \STATE Sample batch of transitions from the replay buffer
                \STATE Update $\theta$ using \cref{eq:gradient} for each sampled transitions, $Q_{\text{lo}}$ stays unchanged
                \STATE set $\theta_- = \theta$ every $T$ steps
             \ENDWHILE                         
        \ENDFOR
	\end{algorithmic}
    \label{alg:deep_corrections}
\end{algorithm}

Instead of learning a Q-function representing the full problem, an additive correction is represented as a neural network. By using this representation of the corrective term, all the innovations (dueling, prioritized replay, double Q-network) used to improve the DQN algorithm can also be used to learn the correction. As illustrated in \cref{alg:deep_corrections}, the pseudo-code of the deep corrections algorithm is very close to the one of DQN with the exception of the representation of the Q function (line 7 and 11). Although the algorithms are similar, learning to represent the correction to an existing value function should be much easier than learning the full value function. In the extreme case where $Q_\text{lo} = Q^*$, $\delta$ should represent a constant function equals to 0. In contrast, if $Q_\text{lo}$ is random, then learning $\delta$ is as hard as learning the value function using standard DQN.

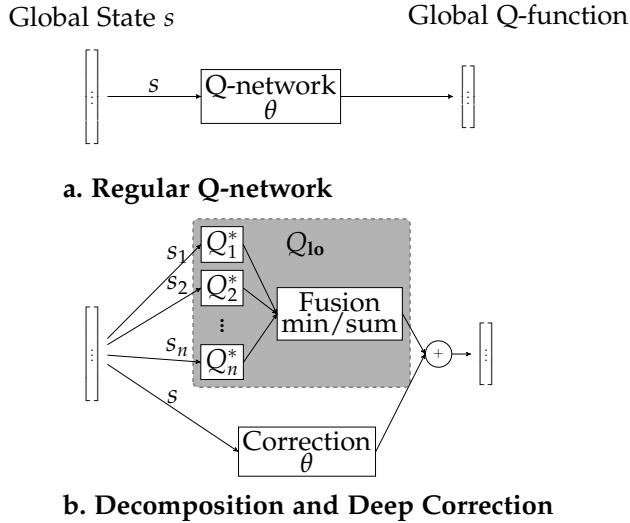
\begin{figure}[t!]
  \centering
	\resizebox {0.6\columnwidth} {!} {
\pgfdeclarelayer{background}
\pgfdeclarelayer{foreground}
\pgfsetlayers{background,main,foreground}

\begin{tikzpicture}[
stop/.style = {regular polygon, regular polygon sides=8,
	draw=red, double, double distance=2mm, ultra thick,
	fill=red, font=\Huge\bfseries, text=white,
	inner sep=0mm, node contents={STOP}}
]


\node (inpt) {$\begin{bmatrix}
	 \\
	 \\
	 \vdots \\
	 \\
	 \\
	\end{bmatrix}$} ;


\node[draw, minimum width=2cm, minimum height=1cm, below right=0.5cm and 3.5cm of inpt, align=center] (correction) {\huge Correction \\ \huge $\theta$};

\node[draw, fill=white, minimum width=1cm, minimum height=0.5cm, above right=1cm and 2.5cm of inpt, align=center] (q1) {\huge $Q^*_1$};

\node[draw, fill=white, minimum width=1cm, minimum height=0.5cm, below=0.2cm of q1, align=center] (q2) {\huge $Q^*_2$};

\node[minimum width=1cm, minimum height=0.5cm, below=0cm of q2, align=center] (qdots) {\huge $\vdots$};

\node[draw, fill=white, minimum width=1cm, minimum height=0.5cm, below=0.2cm of qdots, align=center] (qn) {\huge $Q^*_n$};

\node[draw,fill=white, minimum width=1cm, minimum height=0.5cm, below right=-0.5cm and 0.9cm of q2, align=center] (fusion) {\huge Fusion \\  \huge min/sum};

\node[draw, circle, minimum width=0.5cm, minimum height=0.5cm, right=8.5cm of inpt, align=center] (add) {$+$};

\node (out)[right= 0.5cm of add] {$\begin{bmatrix}
	\\
	\vdots \\
	\\
	\end{bmatrix}$} ;


\draw[->] (inpt) -- node[above]{\huge $s$}(correction.west);

\draw[->] (inpt) -- node[above, near end]{\huge $s_1$} (q1.west);
\draw[->] (inpt) -- node[above=0.3em, near end]{\huge $s_2$}(q2.west);
\draw[->] (inpt) -- node[above, near end]{\huge $s_n$}(qn.west);

\draw[->] (q1.east) -- (fusion.west);
\draw[->] (q2.east) -- (fusion.west);
\draw[->] (qn.east) -- (fusion.west);

\draw[->] (fusion.east) -- (add.west);
\draw[->] (correction.east) -- (add.west);
\draw[->] (add.east) -- (out.west);


\begin{scope}[on background layer]
	\path (qn.west |- qn.south)+(-0.2,-0.2) node (a) {};
	\path (fusion.east |- qn.south)+(0.2,-0.2) node (b) {};
	\path (fusion.east |- q1.north)+(0.2,0.2) node (c) {};
	\path (qn.west |- q1.north)+(-0.2,0.2) node (d) {};
	\draw[fill=black!30!white, rounded corners, dashed, very thick, draw=black!50!white] (a) rectangle (c);
	\node[right=1cm of q1] (qlo) {\huge \textbf{$Q_{\text{lo}}$}};
\end{scope}


\node[above=4cm of inpt] (inptq) {$\begin{bmatrix}
	\\
	\\
	\vdots \\
	\\
	\\
	\end{bmatrix}$};

\node[above =0.3cm of inptq] (inptqlabel) {\huge Global State $s$};

\node[draw, minimum width=2cm, minimum height=1cm, right=2.5cm of inptq, align=center] (qnet) {\huge Q-network \\ \huge $\theta$};

\node (outq)[right= 3cm of qnet] {$\begin{bmatrix}
	\\
	\vdots \\
	\\
	\end{bmatrix}$} ;

\node[right=6cm of inptqlabel] (outqlabel) {\huge Global Q-function};

\draw[->] (inptq.east) -- node[above]{\huge $s$} (qnet.west);
\draw[->] (qnet.east) -- (outq.west);

\node (a) [below left=0.6cm and 0.5cm of inptq, align=left, anchor=north west] {\textbf{\huge a. Regular Q-network}};

\node (b) [below left=2.2cm and 0.5cm of inpt, align=left, anchor=north west] {\textbf{\huge b. Decomposition and Deep Correction}} ;

\end{tikzpicture}

	
	}
	\caption{Regular DQN architecture (a) and architecture of the deep correction approach used with the decomposition method (b). The global state is decomposed and each sub-state is fed into networks pre-trained on the single entity problem. The output of the correction network is then added to the output of the utility fusion to approximate the global Q-function. More generally, the shaded gray area could be replaced by any existing value function $Q_{\text{lo}}$.}
	\label{fig:architecture}
\end{figure}

An illustration of the deep correction architecture along with the decomposition method is presented in \cref{fig:architecture}.
In order to use this approach jointly with a decomposition method, we can substitute $Q_{\text{lo}}$ by the results from utility fusion. For example substituting \cref{eq:sum} in \cref{eq:correction} would lead to the following approximation of the global $Q$ function:
\begin{equation}
		Q^*(s, a) \approx \sum_i Q^*_i(s_i, a) + \delta(s, a; \theta)
		\label{eq:sum-corr}
\end{equation}
The value functions $Q^*_i$ come from the subproblem solutions. In the collision avoidance setting, the value functions are the same, although other settings could use distinct value functions to represent $Q^*_i$. We can see from the update rule with correction that these functions are constant with respect to $\theta$. In practice, if a neural representation is used for the subproblem solutions, then the weights of the subproblem networks are frozen during the training of the correction term. 

One advantage of framing the correction method as a multi-fidelity optimization problem is the flexibility in the method of obtaining $Q_\text{lo}$. There are no particular constraints on the representation of $Q_\text{lo}$. Various approaches can be used to obtain an approximate value function for the problem. We could perform policy evaluation given a rule-based controller or use a model-based approach, for example. The low-fidelity approximation of the value function is a way to introduce prior knowledge in the algorithm. The fidelity of $Q_\text{lo}$ corresponds to how close it approximates the optimal value function $Q^*$. In this work we focused on decomposition methods to provide such an approximation, as they have been shown to provide good approximations for global problems empirically. Those methods often rely on some independence assumptions between the entities~\cite{russel2003, tan1993}. The role of the correction network is to fill the difference between the approximated function and the true value function of the global problem where entities are not independent. Hence $\delta$ plays a more important role the further $Q_\text{lo}$ is from the true value function.


A limitation of our method is that the correction network could potentially get stuck in a local optima. Moreover it could harm the performance during training as it is initialized randomly. Further work must be done to provide constraints on the policy improvement or convergence guarantees.
In this work, we do not provide theoretical guarantees on the convergence of the deep Q-learning algorithm used to learn the correction term. However, we demonstrate in \cref{sec:fisheries} and \cref{sec:crosswalk} that despite of the lack of guarantees, the deep Q-learning approach still learns good policies.  Intuitively, starting with the policy from utility fusion significantly reduces the amount of exploration in DQN. If the low-fidelity value function is a good approximation of the optimal policy, the update rule leads the corrective term towards zero. A regularization term could be added to the loss function to minimize the impact of the corrective factor if one has trust in the low-fidelity policy. Although we used a neural network representation for $\delta$,  the multi-fidelity formulation from \cref{eq:correction} generalizes to any class of parametric functions.

The remaining sections of the paper are dedicated to applying the deep correction approach in addition to the decomposition method. They provide empirical results that demonstrates the advantage of learning the correction term. The problems and the deep correction algorithm were implemented using the POMDPs.jl framework \cite{egorov2017}. An open source implementation of the deep corrections algorithm is available at \url{https://github.com/sisl/DeepCorrections.jl}.

\section{Fisheries Management}\label{sec:fisheries}

This section presents an application of the deep corrections method used in a multi-agent problem. In this example, the goal is to coordinate multiple agents to maximize a global objective. We consider a similar fisheries management problem as \citeauthor{russel2003}~\cite{russel2003}. This problem belongs to the broader category of multi-agent resource allocation problems, which has also been applied to server allocation for different applications \cite{tesauro2005}. These two previous works rely on decomposition methods to find approximately optimal policies.

\begin{figure}[!h]
	\centering 
	\includegraphics[width=0.7\columnwidth]{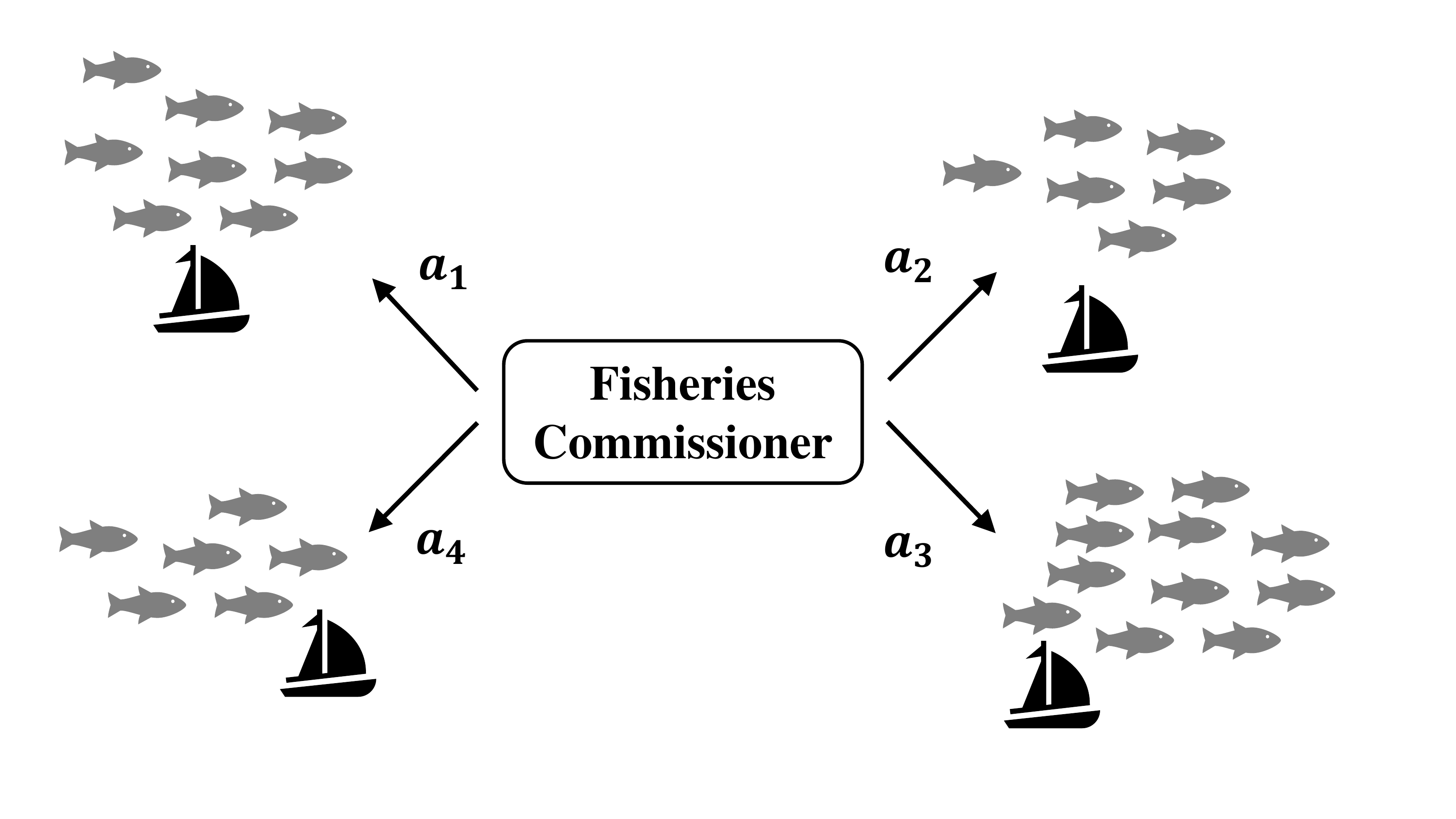}
	\caption{Illustration of the fisheries management problem with four boats. The fisheries commissioner must assign the proportion of fish that each fisherman can fish to maximize the aggregate catch over time.}
\end{figure}

\subsection{Model Description}
 Several fishermen must coordinate to catch a maximum number of fish. If each fisherman follows a selfish strategy, the fish stocks will collapse resulting in a ``tragedy of the commons." The fisheries commissioner, a centralized arbitrator, must assign a proportion of the local fish population that each fishermen is allowed to catch at each season. The goal of the commissioner is to maximize the global catch. We consider a fishery with $n$ fishermen fishing in local regions. A season consists of two steps: a mating season and a fishing season. During the mating season, the fish population reproduces and is then spread across $n$ regions with equal probability. During the fishing season, each fisherman is assigned a proportion of fish to catch in their region. 

The fisheries management problem can be modeled as a multi-agent MDP:
\begin{itemize}
	\item \emph{State space}: the state consists of the number of fish in each region at time $t$: $s_t = (f_{1t}, \ldots, f_{nt})$. The local task of the fisheries problem has a one dimensional state equals to $f_it$ at time $t$ for fisherman $i$. 
	
	\item \emph{Action space}: Let $\mathcal{A}_{\text{local}} = \{1, 0.5, 0.3, 0.1\}$ the possible individual actions, corresponding to a proportion of fish to catch in a given region. The global action space consists of the joint action space, $\mathcal{A} = \mathcal{A}_{\text{local}}^n$. Given $a \in \mathcal{A}_{\text{local}}^n$, $a_i$ corresponds to the proportion of fish assigned to fisherman $i$.
	\item \emph{Transition model}: During the mating season, the fish population reproduces according to the following model: 
	\begin{equation}
	f_{t+1} = f_t \exp (G(1 - \frac{f_t}{f_\text{max}})
	\end{equation}
	where $f_t = \sum_{i=1}^n f_{it}$ is the total fish population at time $t$, $G$ is the population growth rate, and $f_\text{max}$ is the maximum fish population. After the mating season, the fish are uniformly assigned to the different regions. The number of fish, $c_{it}$, caught by the fisherman $i$, follows a Poisson distribution of mean $\eta a_{it} f_{it}$ where $\eta$ is a boat efficiency parameter and $a_{it}$ is the proportion of the local fish population assigned to fisherman $i$ at time $t$.  
	
	\item \emph{Reward model}: The individual reward of fisherman $i$ is given by $r_i(c_{it}, a_{it}) = C(c_{it} - \zeta a_{it}^2)$, where $\zeta$ is a constant reflecting the cost of fishing and $C$ is a normalization factor equal to $n/f_\text{max}$. The term proportional to $a_{it}^2$ reflects the fact that catching more fish requires more resources such as equipment or crew members and is hence more costly. We used the same reward definition as \citeauthor{russel2003}. The global reward consists of the sum of the individual rewards. 
	
\end{itemize}

An episode ends after the fishery has survived for \num{100} seasons or when the fish population goes below a minimum. 
The numerical values for the parameters of the fisheries problem are summarized in \cref{tab:fisheries-param}.

\begin{table}[t]
	\centering
	\caption{Parameters for the fisheries management problem}
	\begin{tabular}{@{}lS@{}} 
		\toprule[1pt]
		Parameter & \text{Value} \\
		\midrule
		number of boats & 10 \\
		initial population & 1.5e5  \\
		maximum population & 3e5 \\
		minimum population & 200 \\
		growth rate & 0.5 \\
		fishing cost $\zeta$ & 1e3 \\
		boat efficiency $\eta$ & 0.98 \\
		discount factor & 0.99 \\    
		\bottomrule[1pt]
	\end{tabular}
	\label{tab:fisheries-param}
\end{table}

We compare the performance of three different policies against simple baselines.

\textbf{Baselines:} The first baseline is a random policy where the commissioner selects a random action for each boat. The other baselines consists of single action policies where each boat is allowed to fish a fixed proportion of fish at every step. The policy with fixed action \num{0.1} (harvesting a tenth of the fish population) corresponds to a conservative policy where each boat fish the minimum possible number of fish. The policy with the fixed action \num{1.0} is a greedy policy.

Solving the full problem using the conventional DQN algorithm is intractable. The size of the action space grows exponentially with the number of boats: with 10 boats, there are $4^{10}$ possible actions. Instead, we use the different decomposition methods presented below.

\textbf{Max-sum:}
We first solve a single agent problem of assigning the proportion of fish to catch for one boat in one region. The environment follows the same model as in the multi-agent setting but with an initial population reduced to \num{15000}. A Q-network is computed using DQN, with a one-dimensional input corresponding to the current amount of fish available. To generate a multi-agent policy, we summed the individual utilities given by the solution to the single-agent problem as follows:
\begin{equation}
	a^* = \argmax_{a_1, \ldots, a_n} \sum_i Q_{\text{single}}(s_i, a_i)
\end{equation}
where $s_i$ is the number of fish in region $i$ after the fishing season and $a_i$ the local action of fisherman $i$.

\textbf{Decomposed DQN:} We learn $n$ different Q-networks in a decomposed manner as suggested by \citeauthor{russel2003} with the difference that each Q-network takes the global state into account and is trained using the global reward signal~\cite{russel2003}. At every training step, we perform one parameter update for each Q-network. Each network predicts the value of the global state for each local action. The dimension of the output is $|\mathcal{A}_{\text{local}}|$. The joint action is again given by maximizing the sum of the utilities:
\begin{equation}
	a^* = \argmax_{a_1, \ldots, a_n} \sum_i Q_{i}(s, a_i; \theta_i)
\end{equation}

\textbf{Max-sum with correction:} Finally, we apply the deep corrections method to improve the max-sum policy. The correction function is represented by a neural network and receives the full state as input but predicts the value of the local action (same input output as for the decomposed DQN). The correction term is trained using a decomposed version of \cref{alg:deep_corrections} (one correction network per boat). The joint policy is then given by:
\begin{equation}
	a^* = \argmax_{a_1, \ldots, a_n} \sum_i Q_{\text{single}}(s_i, a_i) + \delta_i(s, a_i; \theta_i)
\end{equation} 
During the training of the correction terms, the low fidelity approximation given by $Q_{\text{single}}$ is not updated. The correction network has access to more information than the original policy. 

\subsection{Experiments}

In order to fairly compare the three methods, we allocate a fixed training budget of \num{160}k experience samples. For the deep corrections approach, the budget is split as follows: \num{100}k examples are used to learn the single agent Q-network and \num{60}k examples are used to train the correction networks. For the decomposed DQN approach, the entirety of the budget is used to learn the Q-networks. In all three trainings, the same neural network architecture is used: one hidden layer of sixteen nodes with rectified linear units activations. The dimensionality of the input is either one (for the single agent problem) or ten (for the multi-agent problem), and the output of the networks is always of dimension $|\mathcal{A}_{\text{local}}|=4$. Larger architectures did not improve performance or would require more experience samples to achieve a similar accumulated reward. For the three policies, we used the same hyperparameters, shown in \cref{tab:fish_dqnparam}, and an $\epsilon$-greedy policy with a linearly decaying $\epsilon$ for exploration. 

To evaluate the policies, we measure the average accumulated undiscounted reward across \num{100} simulations on the fisheries management problem. The results from these simulations are reported in \cref{fig:fisheries_performance} for the five baselines policies, the max-sum policy, the decomposed DQN policy and the policy improved with deep corrections.

\begin{table}[h]
	\centering
	\caption{Deep Q-learning parameters for the fisheries management problem}
	\begin{tabular}{@{}ll@{}}
		\toprule[1pt]
		Hyperparameter & Value \\
		\midrule
		Neural network architecture & 1 fully connected layer of 16 nodes \\
		Activation functions & Rectified linear units \\
		Replay buffer size & \SI{500}k experience samples \\
		Target network update frequency & \SI{2}k episodes \\
		Discount factor & \SI{0.99}{} \\
		Optimizer & Adam \\
		Learning rate & \SI{1e-4}{} \\
		Prioritized replay \cite{schaul2016}  & $\alpha=$\SI{0.7}{}, $\beta=$\SI{1e-3}{} \\
		Exploration fraction & \SI{0.2}{} \\
		Final $\epsilon$  & \SI{0.05}{}  \\ 
		\bottomrule[1pt]
	\end{tabular}
	\label{tab:fish_dqnparam}
\end{table}

\subsection{Results}

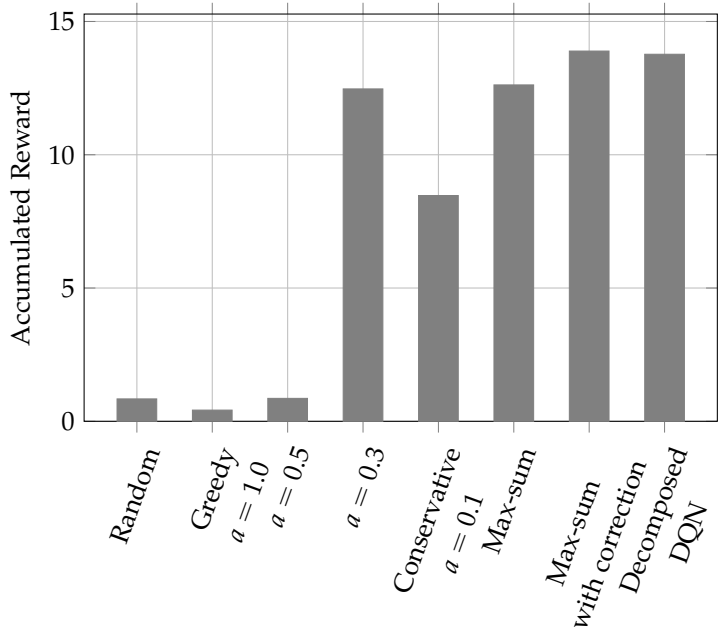
\begin{figure}[h]
	\centering 

\begin{tikzpicture}[]
\begin{axis}[
  grid = both,
  width = 10cm,
  height = 7cm,
  ylabel = {Accumulated Reward}, 
  ybar=0pt, 
  xtick=data, 
  x tick label style = {align = center, rotate = 70, anchor = north east},
  symbolic x coords={Random, Greedy \\ $a=1.0$, $a=0.5$, $a=0.3$, Conservative \\ $a=0.1$, Max-sum, Max-sum \\ with correction, Decomposed \\ DQN},
  bar width=15pt, 
  ymin = {0}
]
\addplot+ [gray]coordinates {
(Random, 0.84)
(Greedy \\ $a=1.0$, 0.42)
($a=0.5$, 0.86)
($a=0.3$, 12.47)
(Conservative \\ $a=0.1$, 8.47)
(Max-sum, 12.62)
(Max-sum \\ with correction, 13.89)
(Decomposed \\ DQN, 13.77)
};
\end{axis}

\end{tikzpicture}
	\caption{Performance of different policies on the fisheries management problem with ten boats.}
	\label{fig:fisheries_performance}
\end{figure}

We can see from \cref{fig:fisheries_performance} that the learned policies outperform all the baselines. As expected, the greedy policy, the random policy, and the $a=0.5$ policy (harvesting half of the population all the time) perform poorly and are victims of ``the tragedy of the commons". All other policies allowed the fisheries to survive for a hundred seasons because the stock never ran out. The conservative policy reaches a score of \num{8.47} and is outperformed by the policy harvesting \num{0.3} of the fish population which scored \num{12.47}. The max-sum approach gives a significant improvement over the conservative policy with a score of \num{12.62}, but only a minimal improvement compared to the the policy with $a=\num{0.3}$. When learning the corrective term in addition to the conservative policy, the score was improved as it achieves \num{13.9}. The decomposed DQN approach achieved a close performance of \num{13.7}, which is lower than the policy with corrections. In this problem, the gap between the correction method and the DQN approach is not very large, but the correction network converged with \SI{60}k samples only (against \SI{160}k for the decomposed DQN). 

These results show that the deep corrections method efficiently improves an existing policy and that utility decomposition is not optimal. Moreover, this result demonstrates that the corrected policy achieves better performance than a policy trained to solve the full problem using DQN, or in this case a decomposed DQN approach.

\section{Occluded Crosswalk}\label{sec:crosswalk}

This section applies the deep correction approach to an autonomous driving scenario where a vehicle must avoid multiple entities. We demonstrate how to apply the decomposition method by considering each entity independently and show that learning the correction term improves the resulting policy.  

\subsection{Model Description}

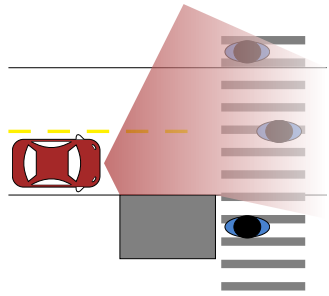
\begin{figure}[t]
	\centering
	\resizebox{0.3\columnwidth}{!}{
		\begin{tikzpicture}[
stop/.style = {regular polygon, regular polygon sides=8,
      draw=red, double, double distance=2mm, ultra thick,
      fill=red, font=\Huge\bfseries, text=white,
      inner sep=0mm, node contents={STOP}}
]

\definecolor{myblue}{RGB}{72,130,206}
\definecolor{myred}{RGB}{165,40,40}

	\draw [very thick] (-5,4) -- (15,4);
	
	\draw [very thick] (-5,-4) -- (15,-4);
	
    \draw [line width=6, color=yellow, dash pattern=on 40pt off 30pt] (-5,0) -- (7,0);

    \draw [line width=150, color=gray,dash pattern=on 15pt off 25pt] (11,-10) -- (11, 6);
    
    \draw [fill=gray] (2,-4) rectangle (8,-8);
    
    \node [inner sep=0pt] (ped1) at (10,-6)
    	{\includestandalone[width=80pt]{pedestrian}};
    \node [inner sep=0pt] (ped2) at (12,0)
       	{\includestandalone[width=80pt]{pedestrian}};
    \node [inner sep=0pt] (ped1) at (10,5)
    {\includestandalone[width=80pt]{pedestrian}};
    
      \shade[top color=myred!40!myred, shading angle=90, opacity=0.65] 
       (1,-2) -- (6,8) -- (15, 4) -- (15, -4) -- (15,-5.5) -- (8,-4) -- (2,-4) --
       cycle;
%
	
	\node [inner sep=0pt] (egoCar) at (-2,-2)
	{\includestandalone[angle=-90]{red_car}};
	
%
%

\end{tikzpicture}
	}
	\caption{The autonomous vehicle in red must choose the acceleration to apply to navigate safely through the crosswalk. A gray obstacle occludes the vehicle's field of view. Several pedestrians might be present for the ego vehicle to avoid.}
	\label{fig:scenario}
\end{figure}

To illustrate the policy correction technique, we use an autonomous driving scenario involving a crosswalk occluded by a physical obstacle as presented in \cref{fig:scenario}. The objective of the agent (the ego car) is to navigate through the crosswalk safely and efficiently. It must avoid potential pedestrians crossing the road. The agent must anticipate uncertainty and trade off between gathering information and reaching the goal position.

This problem can be modeled as a partially observable Markov decision process (POMDP). Contrary to an MDP, the state is not fully observable. In this crosswalk scenario, partial observability comes from the presence of a physical obstacle occluding the field of view of the sensors. Previous works propose modeling tactical decision making for autonomous driving as a POMDP \cite{bai2014, brechtel2014, bandyopadhyay2012}. A major challenge is to find a representation that makes the problem tractable. \citeauthor{bandyopadhyay2012} suggest a discrete formulation to address a pedestrian collision avoidance problem~\cite{bandyopadhyay2012}. Although the method provides an efficient policy, it is unlikely to scale in avoiding multiple pedestrians since the size of the state space would grow exponentially. Other approaches solve the problem in the continuous space and rely on sampling based methods, but suffer from the curse of dimensionality when trying to handle multiple road users \cite{bai2014, brechtel2014}. \citeauthor{wray2017} suggest a decomposition of complex driving scenario into several POMDPs~\cite{wray2017}. Each instantiated problem is solved offline. An arbitrator chooses online between the actions given by the different policies. The arbitration process relies on a simple ordering of the actions and does not take into account the utility of the individual problems. In this work, we suggest a similar formulation of the problem where each pedestrian present in the environment correspond to a different instance of a POMDP. To arbitrate between each entity to avoid, we use the utility fusion method presented in \cref{sec:fusion}. We learn a corrective term to refine the approximation and improve performance. Previous works address an autonomous braking problem at crosswalk using reinforcement learning~\cite{chae2017}. However, they do not address the scalability of the approach in environments with multiple pedestrians to avoid. 

The state of the environment consists of the position and velocity of the ego car as well as the position and velocity of the pedestrians. The autonomous vehicle measures its own position and velocity, the position and velocity of the pedestrians that are not occluded, and a constant value $s_{\text{absent}}$ for the pedestrians that are not visible. Observing $s_{\text{absent}}$ can mean either that a pedestrian is in the occluded area or that it is simply absent from the environment.

The action space consists of a set of strategic maneuvers such as hard breaking, moderate braking, keeping a constant speed, and accelerating, as represented by a finite set of acceleration inputs: 
\[\{\SI{-4}{\meter\per\second\squared},\SI{-2}{\meter\per\second\squared},\SI{0}{\meter\per\second\squared},\SI{2}{\meter\per\second\squared}\} \text{.}\]
A transition model is defined that uses a point mass model for the ego vehicle and a constant speed model with random noise for the pedestrian. When a pedestrian is in the state $s_{\text{absent}}$, it may appear in the environment (at the beginning of the crosswalk) with a probability of \num{0.3}. When present in the environment, the pedestrian follows a desired speed of \SI{1}{\meter\per\second}, At each time steps the pedestrian changes its speed by a random amount in the set: $\{\SI{-1}{\meter\per\second},\SI{0}{\meter\per\second},\SI{1}{\meter\per\second}\}$. The motion model of the pedestrian does not depend on the state of the car. It is designed such that it spans a very large number of possible trajectories. It is expected that the ego vehicle learns a very conservative policy when interacting with this environment. Many of the pedestrian trajectories generated using the proposed motion model would not occur in the real world. A more complex modeling of pedestrian behavior is left as future work. Previous work presents data driven approaches for modeling pedestrian behavior at crosswalk~\cite{zhao2017}.

The agent receives a reward when reaching a terminal state. Terminal states are defined by three possible outcomes:
\begin{itemize}
	\item A collision, for which the agent receives a penalty of $-1$.
	\item A time out if the agent fails to pass the crosswalk under \SI{20}{\second}, for which the reward is $0$.
	\item A success if the agent reaches a goal position about \SI{6}{\meter} after the crosswalk, for which the reward is $+1$.
\end{itemize}
Time efficiency is not explicitly present in the reward function. Instead, a discount factor of $0.99$ will encourage the vehicle to gather the final reward as fast as possible, since the value of the reward decays exponentially with time. Given the structure of the reward function, two evaluation metrics are defined. The first one is the collision rate across many simulations, and the second one is the average time to pass the crosswalk. These metrics account for safety and efficiency respectively. There is a natural trade-off between these two objectives, which makes this autonomous driving scenario a multi-objective optimization problem.
The state variable as formulated consists of two dimensions for the ego vehicle and two dimensions for each pedestrian considered. As a consequence, the size of the state space grows exponentially with the number of agents considered. Many POMDP solvers suffer from this representation \cite{bai2014, bandyopadhyay2012, brechtel2014}. Instead, the computational cost of the decomposition only increases linearly in the number of agents considered. One query of the individual utility functions per pedestrian is required to combine the utilities online.

We compare the performance of three different approaches to solving this occluded crosswalk scenario.

\vspace{0.5cm}

\textbf{Deep Q-Learning:}
The first approach uses deep Q-learning to solve the full problem. This scenario is naturally continuous and can be handled by a neural network representation of the value function. We used the hyperparameters from \cref{tab:dqnparam}. A challenge for the occluded crosswalk problem is that the number of pedestrians present in the scene is unknown. For the sake of simplicity, the number of pedestrians was capped to ten. As a result, there are twenty-two state dimensions, with two dimensions for the ego car position and velocity and two dimensions for each pedestrian position and velocity.

When a pedestrian is not observed, its position and velocity are set to the constant value $s_{\text{absent}}$. To address partial observability in the neural representation we use a $k$-Markov approximation that approximates the POMDP structure as an MDP where the state consists of the last $k$ observations: $s_t = (o_{t-k}, o_{t-k+1}, \cdots, o_t)$. The Q-network is given a history of the last four states~\cite{mnih2015}. This implementation limitation could be improved by using recurrent neural network to handle the partial observability, but this is left as future work.

\vspace{0.5cm}

\textbf{Utility Fusion:}
The occluded crosswalk problem originally requires avoiding multiple pedestrians. A good decomposition of the utility for this problem considers the utility of avoiding a single pedestrian, similar to avoiding a single intruder aircraft \cite{chryssanthacopoulos2012}. For each pedestrian $i$ in the environment, the ego vehicle observes a state $s_i$ consisting of its own position and velocity as well as the position and velocity of the pedestrian. The agent then computes the optimal state-action value $Q^*(s_i, a)$ assuming $i$ is the only user to avoid. The state $s_i$ is four dimensional and consists of the position and velocity of the ego vehicle as well as the position and velocity of the $i$-th pedestrian. The state-action value measures the expected accumulated reward of taking action $a$ and then following the optimal policy associated with user $i$. By using \cref{eq:fusion}, the agent can approximate the solution of the global value function. In this work we considered two functions to combine the value functions as presented in \cref{eq:sum,eq:min}.

\vspace{0.5cm}

\textbf{Policy Correction:}
Using the two policies described above as a low-fidelity approximation of the optimal value function, we learn an additive corrective term. The correction function is represented by a neural network, which is optimized through the approach presented in \cref{sec:correction}. The input to the correction term is similar to the one used for the deep Q-learning approach.

In this collision avoidance problem, the number of pedestrians to avoid is not known in advance due to the presence of the obstacle. To address the issue, we capped the number of possible pedestrians to consider to ten. This maximum number of pedestrians was measured empirically and may change with a different probability of appearance for the pedestrians.  

The source code for the simulation environment is available at \url{https://github.com/sisl/AutomotivePOMDPs.jl}.

\subsection{Experiments}

All proposed solutions methods involve training a Q-network: on the full problem, on the single pedestrian problem, and on the corrective term. In order to fairly compare the three methods, a fixed training budget of one million experience samples was used for all three methods. 
For the policy correction technique, part of the sample budget was used to train the decomposed policy. The sample budget is divided equally between training a Q-network on the single pedestrian problem and training the corrective term. A hyperparameter search was executed to select the neural network architecture and the target network update frequency. Other parameters were set to common values found in the literature \cite{mnih2015, schaul2016, wang2016}.

\begin{table}[t]
	\centering
	\caption{Deep Q-learning parameters for the occluded crosswalk problem}
	\begin{tabular}{@{}ll@{}}
		\toprule[1pt]
		Hyperparameter & Value \\
		\midrule
	    Neural network architecture & 5 fully connected layers of 32 nodes \\
	    Activation functions & Rectified linear units \\
		Replay buffer size & \SI{400}k experience samples \\
		Target network update frequency & \SI{5}k episodes \\
		Discount factor & \SI{0.99}{} \\
		Optimizer & Adam \\
		Learning rate & \SI{1e-4}{} \\
		Prioritized replay \cite{schaul2016}  & $\alpha=$\SI{0.7}{}, $\beta=$\SI{1e-3}{} \\
		Exploration fraction & search between \SI{0.0}{} and \SI{0.9}{} \\
		Final $\epsilon$  & search between \SI{0.0}{} and \SI{0.1}{}  \\ 
		\bottomrule[1pt]
	\end{tabular}
	\label{tab:dqnparam}
\end{table}

During training, the agent interacts with the environment following an $\epsilon$-greedy policy. The value of $\epsilon$ is scheduled to decrease during training, and we found that the amount of exploration greatly influences the final policy. For the three approaches (training from scratch on the full problem, training on the problem with only one pedestrian, and training the correction factor), a random search over the exploration schedule was conducted. The exploration fraction represents the fraction of total training time used to linearly decay $\epsilon$ from 1.0 to its final value. Each training sample is initialized randomly, the ego vehicle starts at the same position with a velocity uniformly sampled between \SI{6}{\meter\per\second} and \SI{8}{\meter\per\second}. The initial number of pedestrians and their positions and velocities is also randomly sampled. From this initial state, the state is updated given the action of the agent and the dynamic model described in \cref{sec:crosswalk}.

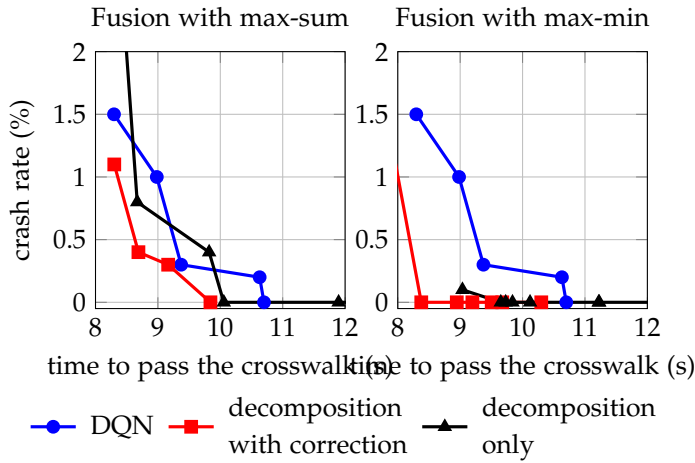
\begin{figure}
	\centering
	\begin{tikzpicture}[]
\begin{groupplot}[
	width=4.9cm,
	height=5cm,
	grid=both,
	group style={horizontal sep = 0.7cm, vertical sep = 1.75cm, group size=2 by 1}]
\nextgroupplot [
	legend columns=3,
	legend style={draw=none,at={(1.0, -0.3)}, 
	column sep=3pt,anchor=north}, 
	ylabel = {crash rate (\%)}, 
	legend style={cells={align=left}},
	title = {Fusion with max-sum}, 
	xmin = {8}, 
	xmax = {12}, 
	ymax = {2.0}, 
	xlabel = {time to pass the crosswalk (s)},
	ymin = {-0.05}, 
]
	
\addplot+ [very thick, mark = {*}, blue, mark options={blue, fill=blue}]coordinates {
(8.2965, 1.5)
(8.9824, 1.0)
(9.3729, 0.30000000000000004)
(10.630500000000001, 0.2)
(10.6999, 0.0)
};
\addlegendentry{DQN}
\addplot+ [very thick, mark = {square*}, red, mark options={red, fill=red}]coordinates {
(8.3008, 1.1)
(8.6879, 0.4)
(9.1644, 0.30000000000000004)
(9.8419, 0.0)
};
\addlegendentry{decomposition \\ with correction}
\addplot+ [very thick, mark = {triangle*}, black, mark options={black, fill=black}]coordinates {
(7.5755, 8.4)
(8.6687, 0.8)
(9.821100000000001, 0.4)
(10.0561, 0.0)
(11.896, 0.0)
(12.302200000000001, 0.0)
(12.6978, 0.0)
};
\addlegendentry{decomposition \\ only}
\nextgroupplot [
	title = {Fusion with max-min}, 
	xmin = {8}, xmax = {12}, 
	ymax = {2.0}, 
	xlabel = {time to pass the crosswalk (s)}, 
	ymin = {-0.05}
]
	
\addplot+ [very thick, mark = {*}, blue, mark options={blue, fill=blue}]coordinates {
(8.2965, 1.5)
(8.9824, 1.0)
(9.3729, 0.30000000000000004)
(10.630500000000001, 0.2)
(10.6999, 0.0)
};
\addplot+ [very thick, mark = {square*}, red, mark options={red, fill=red}]coordinates {
(6.6993, 4.5)
(8.3767, 0.0)
(8.9458, 0.0)
(9.1959, 0.0)
(9.5052, 0.0)
(9.6744, 0.0)
(10.299800000000001, 0.0)
};
\addplot+ [very thick, mark = {triangle*}, black, mark options={black, fill=black}]coordinates {
(9.036700000000002, 0.1)
(9.6511, 0.0)
(9.729000000000001, 0.0)
(9.838500000000002, 0.0)
(10.1198, 0.0)
(11.219700000000001, 0.0)
(11.2288, 0.0)
(13.456000000000001, 0.0)
(16.2593, 0.0)
(18.6143, 0.0)
(23.6322, 0.0)
(30.1, 0.0)
(30.1, 0.0)
};
\end{groupplot}

\end{tikzpicture}
	\caption{Pareto frontiers for the different approaches used to solve the crosswalk problem: DQN, utility decomposition, and utility decomposition with correction. Each point results from evaluating a policy generated by a given set of hyperparameters. The figure is zoomed in the region of optimality (bottom left). The two figures represents the max-sum (left) and the max-min (right) fusion techniques. }
	\label{fig:pareto_front}
\end{figure}

All trained policies are evaluated in an evaluation environment that differs from the training environment. It has a finer time discretization, and the model followed by the pedestrian is less stochastic, as reported in \cref{tab:param}. The weights of the networks are frozen during the evaluation. Each simulation is randomly initialized: random initial velocity for the ego vehicle, random number of pedestrians present with a random position and velocity. Each trained policy is evaluated on the two metrics of interest that are averaged over a thousand simulations: collision rate and time to pass the crosswalk. Since these two objectives are conflicting, Pareto optimality is used to decide if a policy is better than another. Formally, we say that a policy dominates another if it outperforms the other policy in one objective and is no worse in the other objective. From this definition, we can draw the Pareto frontier associated with all the generated policies from the hyperparameter search. The frontiers for the three approaches is represented in \cref{fig:pareto_front}.

\begin{table}[t]
	\centering
	\caption{Environment Parameters}
	\begin{tabular}{@{}lll@{}}
		\toprule[1pt]
		 & \multicolumn{2}{c}{Value} \\
		\cmidrule{2-3}
		Parameter & Evaluation & Training  \\ \midrule
		Position sensor standard deviation & \SI{0.5}{\meter}  & \SI{0.5}{\meter}\\
		Velocity sensor standard deviation & \SI{0.5}{\meter\per\second} & \SI{0.5}{\meter\per\second} \\
		Decision frequency  & \SI{0.5}{\second} & \SI{0.5}{\second} \\
		Simulation time step  & \SI{0.1}{\second} & \SI{0.5}{\second} \\
		Pedestrian velocity maximum noise & \SI{0.5}{\meter\per\second} & \SI{1.0}{\meter\per\second} \\
		\bottomrule[1pt]
	\end{tabular}
	\label{tab:param}
\end{table}

\begin{figure}[h]
	\centering
	\begin{tikzpicture}[]
\begin{groupplot}[
grid=both,
width=4.9cm,
height=5cm,
group style={horizontal sep = 1.0cm, vertical sep = 1.75cm, group size=2 by 2}
]
\nextgroupplot [
legend style={draw=none,at={(1.7,-0.5)}, 
	column sep=3pt,anchor=north}, 
title = {Crashes (\SI{}{\percent})}, 
xlabel = {Number of episodes}, 
x tick scale label style={at={(xticklabel cs:1)},anchor=south west}
]
\addplot+ [thick, mark repeat={3},mark size = {1.0}, blue]coordinates {
	(1.0, 27.700000000000003)
	(20001.0, 27.1)
	(40001.0, 30.200000000000003)
	(60001.0, 29.0)
	(80001.0, 25.900000000000002)
	(100001.0, 27.1)
	(120001.0, 27.5)
	(140001.0, 25.8)
	(160001.0, 27.0)
	(180001.0, 25.900000000000002)
	(200001.0, 12.9)
	(220001.0, 11.700000000000001)
	(240001.0, 14.5)
	(260001.0, 8.0)
	(280001.0, 7.1000000000000005)
	(300001.0, 8.4)
	(320001.0, 7.4)
	(340001.0, 1.1)
	(360001.0, 1.4000000000000001)
	(380001.0, 2.5)
	(400001.0, 1.4000000000000001)
	(420001.0, 5.0)
	(440001.0, 0.2)
	(460001.0, 0.7000000000000001)
	(480001.0, 2.3000000000000003)
	(500001.0, 0.4)
	(520001.0, 4.0)
	(540001.0, 0.7000000000000001)
	(560001.0, 0.2)
	(580001.0, 0.30000000000000004)
	(600001.0, 0.5)
	(620001.0, 3.7)
	(640001.0, 2.1)
	(660001.0, 0.1)
	(680001.0, 0.30000000000000004)
	(700001.0, 0.0)
	(720001.0, 1.1)
	(740001.0, 0.6000000000000001)
	(760001.0, 1.2000000000000002)
	(780001.0, 0.2)
	(800001.0, 1.0)
	(820001.0, 0.6000000000000001)
	(840001.0, 9.700000000000001)
	(860001.0, 10.100000000000001)
	(880001.0, 2.4000000000000004)
	(900001.0, 0.1)
	(920001.0, 0.1)
	(940001.0, 4.800000000000001)
	(960001.0, 12.4)
	(980001.0, 1.7000000000000002)
};
\addlegendentry{DQN}
\addplot+ [thick, mark repeat={3},mark size = {1.0}, red]coordinates {
	(1.0, 27.0)
	(20001.0, 19.700000000000003)
	(40001.0, 10.600000000000001)
	(60001.0, 11.4)
	(80001.0, 1.9000000000000001)
	(100001.0, 0.4)
	(120001.0, 0.6000000000000001)
	(140001.0, 0.30000000000000004)
	(160001.0, 0.30000000000000004)
	(180001.0, 0.30000000000000004)
	(200001.0, 0.4)
	(220001.0, 0.8)
	(240001.0, 0.30000000000000004)
	(260001.0, 0.30000000000000004)
	(280001.0, 0.1)
	(300001.0, 0.0)
	(320001.0, 0.30000000000000004)
	(340001.0, 0.4)
	(360001.0, 0.7000000000000001)
	(380001.0, 0.6000000000000001)
	(400001.0, 0.30000000000000004)
	(420001.0, 1.8)
	(440001.0, 0.4)
	(460001.0, 0.0)
	(480001.0, 0.0)
};
\addlegendentry{Max-min correction}
\addplot+ [thick, mark repeat={3},mark size = {1.0}, black, mark={triangle*}]coordinates {
	(1.0, 29.400000000000002)
	(20001.0, 9.5)
	(40001.0, 5.0)
	(60001.0, 0.1)
	(80001.0, 0.0)
	(100001.0, 0.0)
	(120001.0, 0.0)
	(140001.0, 0.30000000000000004)
	(160001.0, 1.0)
	(180001.0, 0.7000000000000001)
	(200001.0, 5.1000000000000005)
	(220001.0, 6.2)
	(240001.0, 0.6000000000000001)
	(260001.0, 3.6)
	(280001.0, 1.6)
	(300001.0, 0.8)
	(320001.0, 0.2)
	(340001.0, 0.4)
	(360001.0, 4.4)
	(380001.0, 5.300000000000001)
	(400001.0, 0.6000000000000001)
	(420001.0, 0.4)
	(440001.0, 0.2)
	(460001.0, 0.6000000000000001)
	(480001.0, 0.1)
};
\addlegendentry{Max-sum correction}

\nextgroupplot [
title = {Successes (\SI{}{\percent})}, 
xlabel = {Number of episodes},
x tick scale label style={at={(xticklabel cs:1)},anchor=south west}]

\addplot+ [thick, mark repeat={3},mark size = {1.0}, blue]coordinates {
	(1.0, 47.300000000000004)
	(20001.0, 44.800000000000004)
	(40001.0, 58.900000000000006)
	(60001.0, 62.6)
	(80001.0, 67.5)
	(100001.0, 70.5)
	(120001.0, 70.2)
	(140001.0, 72.5)
	(160001.0, 72.3)
	(180001.0, 69.10000000000001)
	(200001.0, 69.5)
	(220001.0, 82.2)
	(240001.0, 81.10000000000001)
	(260001.0, 21.200000000000003)
	(280001.0, 92.4)
	(300001.0, 91.60000000000001)
	(320001.0, 92.60000000000001)
	(340001.0, 98.10000000000001)
	(360001.0, 98.60000000000001)
	(380001.0, 96.4)
	(400001.0, 98.10000000000001)
	(420001.0, 94.5)
	(440001.0, 83.7)
	(460001.0, 95.7)
	(480001.0, 97.7)
	(500001.0, 99.60000000000001)
	(520001.0, 96.0)
	(540001.0, 99.30000000000001)
	(560001.0, 99.7)
	(580001.0, 97.5)
	(600001.0, 99.5)
	(620001.0, 96.30000000000001)
	(640001.0, 97.9)
	(660001.0, 99.9)
	(680001.0, 97.7)
	(700001.0, 99.2)
	(720001.0, 98.9)
	(740001.0, 99.4)
	(760001.0, 98.80000000000001)
	(780001.0, 99.80000000000001)
	(800001.0, 98.80000000000001)
	(820001.0, 99.4)
	(840001.0, 90.30000000000001)
	(860001.0, 89.9)
	(880001.0, 97.60000000000001)
	(900001.0, 99.9)
	(920001.0, 99.9)
	(940001.0, 95.2)
	(960001.0, 87.60000000000001)
	(980001.0, 98.30000000000001)
};
\addplot+ [thick, mark repeat={3},mark size = {1.0}, red]coordinates {
	(1.0, 47.400000000000006)
	(20001.0, 23.8)
	(40001.0, 7.7)
	(60001.0, 56.900000000000006)
	(80001.0, 7.1000000000000005)
	(100001.0, 0.8)
	(120001.0, 0.8)
	(140001.0, 0.4)
	(160001.0, 8.3)
	(180001.0, 5.300000000000001)
	(200001.0, 3.3000000000000003)
	(220001.0, 43.300000000000004)
	(240001.0, 36.7)
	(260001.0, 96.60000000000001)
	(280001.0, 99.4)
	(300001.0, 100.0)
	(320001.0, 99.5)
	(340001.0, 99.60000000000001)
	(360001.0, 99.30000000000001)
	(380001.0, 99.0)
	(400001.0, 99.7)
	(420001.0, 98.2)
	(440001.0, 99.60000000000001)
	(460001.0, 100.0)
	(480001.0, 100.0)
};
\addplot+ [thick, mark repeat={3},mark size = {1.0}, black, mark={triangle*}]coordinates {
	(1.0, 47.1)
	(20001.0, 90.5)
	(40001.0, 95.0)
	(60001.0, 0.2)
	(80001.0, 0.1)
	(100001.0, 0.0)
	(120001.0, 99.9)
	(140001.0, 99.7)
	(160001.0, 99.0)
	(180001.0, 98.80000000000001)
	(200001.0, 94.9)
	(220001.0, 81.0)
	(240001.0, 99.4)
	(260001.0, 96.4)
	(280001.0, 98.4)
	(300001.0, 99.2)
	(320001.0, 99.80000000000001)
	(340001.0, 99.60000000000001)
	(360001.0, 95.60000000000001)
	(380001.0, 94.7)
	(400001.0, 99.4)
	(420001.0, 62.0)
	(440001.0, 99.80000000000001)
	(460001.0, 99.2)
	(480001.0, 99.9)
};
\nextgroupplot [
title = {Time outs (\SI{}{\percent})}, 
xlabel = {Number of episodes},
x tick scale label style={at={(xticklabel cs:1)},anchor=south west}]
\addplot+ [thick, mark repeat={3},mark size = {1.0}, blue]coordinates {
	(1.0, 25.0)
	(20001.0, 28.1)
	(40001.0, 10.9)
	(60001.0, 8.4)
	(80001.0, 6.6000000000000005)
	(100001.0, 2.4000000000000004)
	(120001.0, 2.3000000000000003)
	(140001.0, 1.7000000000000002)
	(160001.0, 0.7000000000000001)
	(180001.0, 5.0)
	(200001.0, 17.6)
	(220001.0, 6.1000000000000005)
	(240001.0, 4.4)
	(260001.0, 70.8)
	(280001.0, 0.5)
	(300001.0, 0.0)
	(320001.0, 0.0)
	(340001.0, 0.8)
	(360001.0, 0.0)
	(380001.0, 1.1)
	(400001.0, 0.5)
	(420001.0, 0.5)
	(440001.0, 16.1)
	(460001.0, 3.6)
	(480001.0, 0.0)
	(500001.0, 0.0)
	(520001.0, 0.0)
	(540001.0, 0.0)
	(560001.0, 0.1)
	(580001.0, 2.2)
	(600001.0, 0.0)
	(620001.0, 0.0)
	(640001.0, 0.0)
	(660001.0, 0.0)
	(680001.0, 2.0)
	(700001.0, 0.8)
	(720001.0, 0.0)
	(740001.0, 0.0)
	(760001.0, 0.0)
	(780001.0, 0.0)
	(800001.0, 0.2)
	(820001.0, 0.0)
	(840001.0, 0.0)
	(860001.0, 0.0)
	(880001.0, 0.0)
	(900001.0, 0.0)
	(920001.0, 0.0)
	(940001.0, 0.0)
	(960001.0, 0.0)
	(980001.0, 0.0)
};
\addplot+ [thick, mark repeat={3},mark size = {1.0}, red]coordinates {
	(1.0, 25.6)
	(20001.0, 56.5)
	(40001.0, 81.7)
	(60001.0, 31.700000000000003)
	(80001.0, 91.0)
	(100001.0, 98.80000000000001)
	(120001.0, 98.60000000000001)
	(140001.0, 99.30000000000001)
	(160001.0, 91.4)
	(180001.0, 94.4)
	(200001.0, 96.30000000000001)
	(220001.0, 55.900000000000006)
	(240001.0, 63.0)
	(260001.0, 3.1)
	(280001.0, 0.5)
	(300001.0, 0.0)
	(320001.0, 0.2)
	(340001.0, 0.0)
	(360001.0, 0.0)
	(380001.0, 0.4)
	(400001.0, 0.0)
	(420001.0, 0.0)
	(440001.0, 0.0)
	(460001.0, 0.0)
	(480001.0, 0.0)
};
\addplot+ [thick, mark repeat={3},mark size = {1.0}, black, mark={triangle*}]coordinates {
	(1.0, 23.5)
	(20001.0, 0.0)
	(40001.0, 0.0)
	(60001.0, 99.7)
	(80001.0, 99.9)
	(100001.0, 100.0)
	(120001.0, 0.1)
	(140001.0, 0.0)
	(160001.0, 0.0)
	(180001.0, 0.5)
	(200001.0, 0.0)
	(220001.0, 12.8)
	(240001.0, 0.0)
	(260001.0, 0.0)
	(280001.0, 0.0)
	(300001.0, 0.0)
	(320001.0, 0.0)
	(340001.0, 0.0)
	(360001.0, 0.0)
	(380001.0, 0.0)
	(400001.0, 0.0)
	(420001.0, 37.6)
	(440001.0, 0.0)
	(460001.0, 0.2)
	(480001.0, 0.0)
};
\end{groupplot}

\end{tikzpicture}
	\caption{Evaluation of the policy performance throughout training. The correction function is being trained on half as many samples as the regular DQN policy but still converges and outperforms the regular DQN policy for both choices of the decomposition method (max-sum or max-min).}
	\label{fig:convergence}
\end{figure}
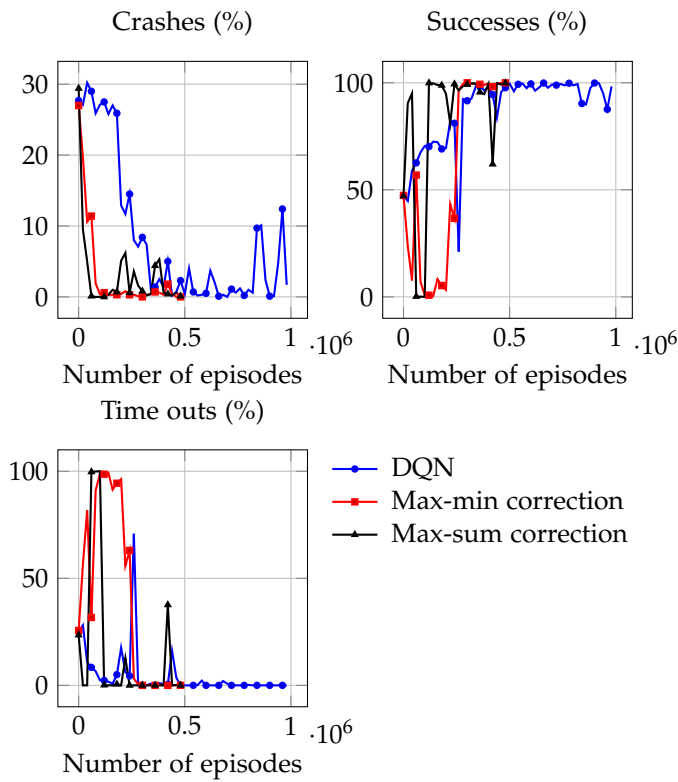

We analyzed the benefit of the policy correction technique in terms of learning speed. Assuming that solving the single pedestrian problem does not require any training, we looked at the evolution of the performance of the corrected policy during training of the correction function. In our application, the single agent policy also requires training a deep Q-network. However, this solution could come from an existing controller or from a model based planning approach that does not require sampling. In \cref{fig:convergence}, we froze the weights of the network at regular intervals during the training and evaluated the policy in the evaluation environment. The corresponding networks in \cref{fig:pareto_front} are the fastest one guaranteeing safety (zero collisions over the thousand simulations). The metrics chosen for this experiments where the number of crashes, successes or time-outs which are the three possible outcomes of a simulation in the occluded crosswalk scenario. A simulation ends in a time-out if the ego vehicle is not able to cross in less than \SI{20}{\second}.

\subsection{Optimality of the Policies} 

The Pareto frontiers in \cref{fig:pareto_front} show a domination of the policy correction method. If we keep one objective fixed, we can always find a policy computed with the correction method that outperforms the two other methods in the other objective. A reasonable approach would be to fix the safety level at \SI{0}{\percent} of collisions over the thousand simulations and use the policy that minimizes the time to pass the crosswalk. Using the max-sum fusion, the fastest safe policy has an average time to cross of \SI{10.06}{\second}. By adding the correction term, the policy achieves an average time to cross of \SI{9.84}{\second}. Using max-min, the policy reaches an average time of \SI{9.65}{\second}, and \SI{8.38}{\second} with the correction. The deep Q-network policy has a time to cross of \SI{10.70}{\second}. For both utility fusion methods, max-sum and max-min, the addition of the corrective factor not only leads to an improvement in the policy but also outperforms the deep Q-network policy. The max-min decomposition method even dominates the Q-network trained on the complex environment. This result illustrates that the choice of a good function for combining the utilities can provide a good policy without training in the multi-pedestrian environment. However, \cref{fig:pareto_front} shows that the max-sum decomposition without correction does not dominate the Q-network approach. In the scenario of interest, considering the agent with the minimum utilities favors risk averse behavior, resulting in safer policies for a given time to pass the crosswalk. 

\begin{figure}[ht]
	\centering
	\input{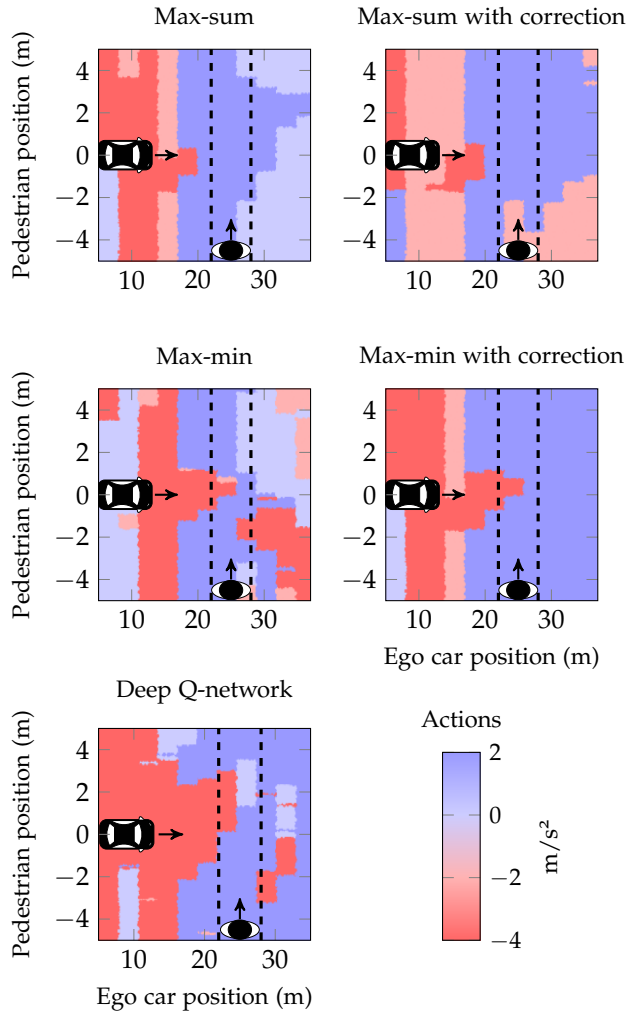}
	\caption{Visualization of the policies from the decomposition methods, the policy correction approach and using a deep Q-network. It assumes that the velocity of the car is \SI{6}{\meter\per\second} and that the pedestrian is not moving. The color at a given point represents the action the ego car would take if it is at a certain position along the road and the pedestrian is at a certain position along the crosswalk.}
	\label{fig:policy_plot}
\end{figure}

To gain intuition on the difference between these policies, we can visualize them on a two-dimensional slice of the state space in \cref{fig:policy_plot}. 
To generate this representation we ran simulations with the ego vehicle driving at a constant speed of \SI{6.0}{\meter\per\second} (not reacting to the policy), and a single pedestrian fixed at a given position along the crosswalk. Although the policy is being tested against one pedestrian, it is still assuming that there might be multiple pedestrians.
\Cref{fig:policy_plot} shows the actions returned by the policy at a given ego car position along the road and a given position of the pedestrian. The pedestrian is located at \SI{25}{\meter} along the road and the ego car is moving along a horizontal line that intersects the crosswalk at \SI{0}{\meter}. All plots show a red zone just on the left of the position $(25, 0)$ representing a braking behavior when a pedestrian is in the middle of the road. The larger to the left the red area is, the more the car will anticipate the braking. Similarly, all the policies present a vertical red zone before \SI{15}{\meter} along the road, indicating that the car will slow down regardless of pedestrian position. This behavior is natural since the presence of the obstacle hides a lot of information. A safe behavior is to slow down until the vehicle has more visibility. These plots can help us visualize areas of the state space where the policy might be suboptimal. For example, the two policies using the decomposition method without the correction present a braking area after the crosswalk. The suboptimal areas after the crosswalk illustrate that all methods used in this paper only approximate the optimal value function. In the area after the crosswalk, braking areas should disappear with more training samples. It is also important to note that the action values in those parts of the state space are very close since it is possible to reach the goal with a probability of 1 in only a couple time steps. These suboptimal parts disappear for the max-min policy with the correction term. In the max-sum case, we can see that the correction relaxed a large part of the braking area to use moderate braking (\SI{-2}{\meter\per\second\squared}) rather than strong braking (\SI{-4}{\meter\per\second\squared}).

\begin{table}[t]
	\centering
	\caption{Best Hyperparameters}
		\begin{tabular}{@{}lSS@{}}
			\toprule[1pt]
			Model & \text{Exploration Fraction} & \text{Final $\epsilon$} \\
			\midrule
			DQN & 0.50 & 0.01\\
			max-min correction & 0.00 & 0.01 \\
			max-sum correction & 0.20 & 0.00 \\
			\bottomrule[1pt]
		\end{tabular}
	\label{tab:bestparam}
\end{table}

\subsection{Training Performance}

We looked at the evolution of the policies performance during training.  Although they were trained with half the samples, the policies with the correction term converge. Since the policy resulting from the decomposition method already has performances close to the deep Q-network policy, only very little exploration is required to learn the corrective term. The max-min decomposition results in more stable training than the other policies, and the three metrics converge after about three hundred thousands episodes. The networks achieving these performances result from the hyperparameter search on the exploration schedule. The corresponding values are reported in \cref{tab:bestparam}. When training the corrective term, the amount of exploration required for good performance is an order of magnitude lower than training DQN on the global problem.

To further analyze the advantages of the deep corrections method for learning efficient policies, we compared the performance of the deep corrections, DQN, and the value decomposition network (VDN) approach proposed by \citeauthor{sunehag2018}\cite{sunehag2018}. For the deep correction method, only \num{50000} training samples are used to train the correction network while the remaining of the training budget is used to pre-train the policy resulting from the decomposition method. We can see that for low training budget, our approach and VDN significantly outperform standard DQN. Moreover the deep corrections algorithms outperforms VDN in most cases. In contrast with VDN, the deep corrections network does not assume any particular structure and should be more expressive. As the training budget increases, all the methods approach the optimal solution. There is also a smaller standard error when using the correction network than when using any of the other approaches. The source code to reproduce these experiments is available at \url{https://github.com/MaximeBouton/DeepCorrectionsExperiments.jl}.

\begin{figure}
    \centering
    \input{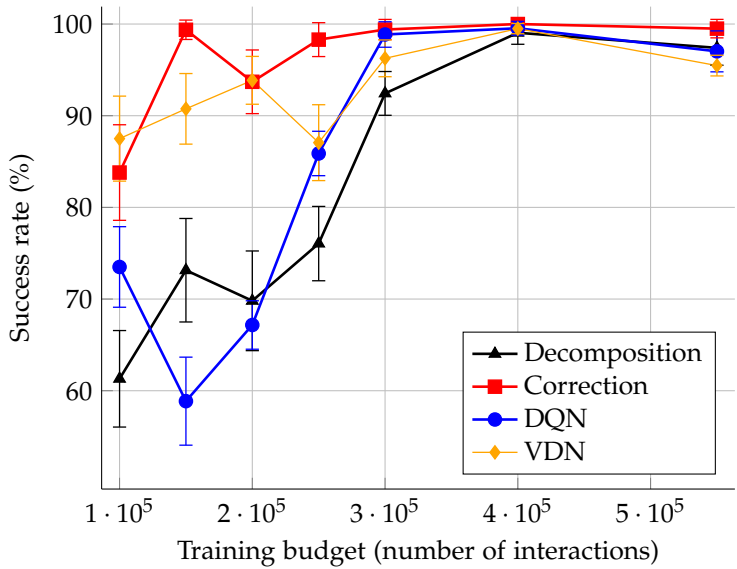}
    \caption{Performance of the deep corrections method, standard DQN and the value decomposition network approach~\cite{sunehag2018} for different training budget. Each point is averaged over five trained policies initialized using different random seeds. The bars represent the standard error. The success rate is computed performing \num{1000} simulations with randomized initial conditions.}
    \label{fig:perf_budget}
\end{figure}


\section{Conclusions and Future work} 

Utility fusion methods efficiently find approximate solutions to decision making problems in multiple agents problems or when a single agent interacts with multiple entities. The global problem is decomposed into local problems involving a single agent interacting with a single entity. Once a solution for the local problems has been computed, solutions can be combined to solve the global problem. Although computationally efficient, combining the utilities with simple functions, such as max-sum and max-min, leads to a suboptimal solution. To overcome this problem, we presented a novel technique to gear an existing suboptimal policy towards the optimum by learning a correction term. The correction term is represented by a neural network and is learned using deep Q-learning. This method inspired from multi-fidelity optimization can significantly improve a policy computed with the decomposition method. We verified this statement empirically on a fisheries management problem and an autonomous driving scenario involving an occluded crosswalk. Adding the corrective factor leads to a much more efficient policy than using the decomposition only. Our method outperformed a deep Q-network policy trained on the full scale problem. We also demonstrated that learning the corrective factor can be done with far fewer training samples than directly learning the value function of the full scale problem.

In the future, more sophisticated multi-fidelity optimization techniques could be used to represent the correction term. Rather than an additive correction, we could try a multiplicative term \cite{eldred2006}. Another possibility involves learning the function used for utility fusion itself~\cite{sunehag2018}. A straightforward extension would be to learn a linear weighted combination of the utilities from the single entity problem. Our approach could be used jointly with the VDN technique where the VDN solution serves as a low fidelity approximation that is being corrected.

Finally, we would like to explore the generality of the correction method. We wish to extend the use of decomposition methods to correcting policies coming from potentially different solving techniques such as an offline POMDP solver. Further experiments for different applications than those presented in this paper could also highlight the benefit of learning a correction to an existing policy.


\printbibliography 

\end{document}